\documentclass[letterpaper]{article} 
\usepackage{aaai2026}  
\usepackage{times}  
\usepackage{helvet}  
\usepackage{courier}  
\usepackage[hyphens]{url}  
\usepackage{graphicx} 
\usepackage{natbib}  
\usepackage{caption} 
\usepackage{float}

\usepackage{multirow}
\usepackage{latexsym}
\usepackage{amssymb}
\usepackage{amsmath}
\usepackage{amsthm}
\usepackage{booktabs}
\usepackage{enumitem}
\usepackage{graphicx}
\usepackage{color}
\usepackage{algorithmicx}
\usepackage{algpseudocode}
\usepackage{booktabs} 
\usepackage{multirow} 
\usepackage{booktabs,multirow,caption,adjustbox}
\usepackage[linesnumbered,ruled,vlined]{algorithm2e}
\usepackage{newfloat}
\usepackage{listings}
\DeclareCaptionStyle{ruled}{labelfont=normalfont,labelsep=colon,strut=off} 
\floatstyle{ruled}
\newfloat{listing}{tb}{lst}{}
\floatname{listing}{Listing}

\begin{document}

\setcounter{secnumdepth}{0} 

\title{TreeFedDG: Alleviating Global Drift in Federated Domain Generalization for Medical Image Segmentation}
\author {
    Yucheng Song\textsuperscript{\rm 1},
    Chenxi Li\textsuperscript{\rm 2,\rm 3},
    Haokang Ding\textsuperscript{\rm 1},
    Zhining Liao\textsuperscript{\rm 4},
    Zhifang Liao\textsuperscript{\rm 1},
}
\affiliations {
    \textsuperscript{\rm 1}School of Computer Science and Engineering, Central South University, Changsha 410083, China\\
     \textsuperscript{\rm 2}College of Computer Science and Technology, Zhejiang University, Hangzhou 310027, China\\
     \textsuperscript{\rm 3}Shanghai Artificial Intelligence Laboratory\\
    \textsuperscript{\rm 4}Glasgow Lab for Data Science \& AI, Public Health, School of Health \& Wellbeing, University of Glasgow, Glasgow, UK\\
}

\maketitle

\begin{abstract}
In medical image segmentation tasks, Domain Generalization (DG) under the Federated Learning (FL) framework is crucial for addressing challenges related to privacy protection and data heterogeneity. However, traditional federated learning methods fail to account for the imbalance in information aggregation across clients in cross-domain scenarios, leading to the Global Drift (GD) problem and a consequent decline in model generalization performance. This motivates us to delve deeper and define a new critical issue: \textbf{global drift in federated domain generalization for medical imaging (FedDG-GD)}. In this paper, we propose a novel tree topology framework called TreeFedDG. First, starting from the distributed characteristics of medical images, we design a hierarchical parameter aggregation method based on a tree-structured topology to suppress deviations in the global model direction. Second, we introduce a parameter difference-based style mixing method (FedStyle), which enforces mixing among clients with maximum parameter differences to enhance robustness against drift. Third, we develop a a progressive personalized fusion strategy during model distribution, ensuring a balance between knowledge transfer and personalized features. Finally, during the inference phase, we use feature similarity to guide the retrieval of the most relevant model chain from the tree structure for ensemble decision-making, thereby fully leveraging the advantages of hierarchical knowledge. We conducted extensive experiments on two publicly available datasets. The results demonstrate that our method outperforms other state-of-the-art domain generalization approaches in these challenging tasks and achieves better balance in cross-domain performance.
\end{abstract}

\section{Introduction}
\begin{figure}[ht]
  \centering
  \includegraphics[width=\linewidth]{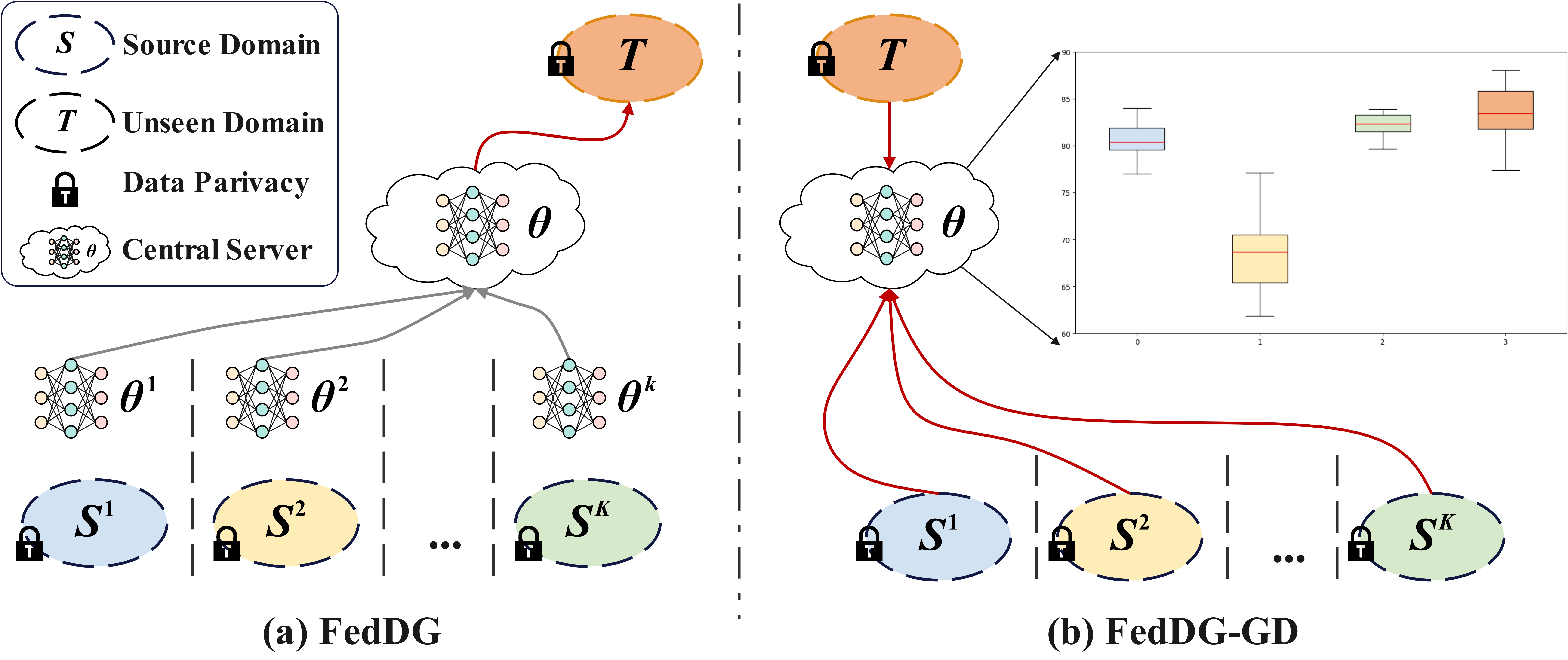}
  \caption{Our Motivation. (a) The basic setup of Federated Domain Generalization (FedDG). (b) When domain heterogeneity is large, the FedDG model suffers from a global drift problem (FedDG-GD), resulting in significant performance differences across different sites.}
  \label{fig1}
\end{figure}

In the field of medical image segmentation, constructing accurate and robust data-driven deep networks necessitates the integration of rich data from multiple medical institutions. However, this demand for cross-institutional collaboration in medical imaging conflicts with the stringent requirements for patient privacy protection \cite{guan2024federated}. In this context, Federated Domain Generalization (FedDG) technology has emerged and is rapidly evolving. This approach builds a “zero-shot” generalization model through distributed training while preserving local data, enabling adaptation to unseen medical domains without requiring additional data during the inference phase \cite{yoon2024domain,zhang2023federated}, as illustrated in Figure \ref{fig1} (a).

Current advancements in FedDG for medical image segmentation tasks have demonstrated promising pilot results. For example, through frequency-space scenario learning \cite{liu2021feddg} or model-level attention combined with style normalization \cite{zhu2023mla}, significant improvements in the model’s adaptability to unseen domains have been achieved. Existing methods often optimize models or align features from multi-source data \cite{wei2024multi,hu2022domain}, resulting in a globally aggregated model with strong generalization capabilities. However, these approaches largely rely on an implicit assumption that, as long as the source domain models are well-trained or features are properly aligned, the aggregation process itself will naturally lead to an optimal global model for unseen domain generalization, with little attention paid to potential biases introduced by the aggregation behavior, especially when inter-domain heterogeneity is substantial. In clinical reality, medical imaging data exhibits multi-domain distribution characteristics due to differences in imaging equipment across institutions (e.g., 1.5T vs. 3.0T MRI), acquisition protocols (e.g., thin-slice CT vs. enhanced CT), and patient populations (e.g., regional pathological variations), leading to significant inter-domain heterogeneity in data distribution. We have found that in medical scenarios, where capturing fine pathological and structural features is essential, the global drift problem is not only prevalent but also has more severe impacts. This causes existing methods to result in the global model deviating from the optimal direction during parameter aggregation, leading to model global drift, as illustrated in Figure \ref{fig1} (b). Theoretically, federated personalization methods could be employed to address the global drift issue, as they allow clients to retain certain personalized features to adapt to local data distributions. However, federated personalization primarily focuses on enhancing the performance of local models in their respective source domains and does not consider how to enable effective generalization of the global model to unseen domains. Therefore, this motivates us to delve deeper and define a new critical problem: global drift in federated domain generalization for medical imaging (FedDG-GD).

Our FedDG-GD problem is clinically significant in medical scenarios due to distribution differences caused by variations in equipment, protocols, and patient populations across hospitals. During parameter aggregation, federated models often deviate from the ideal direction, hindering effective generalization to unseen domains. In particular, in the field of medical imaging, where models need to capture fine pathological or structural features, even minor global drift can result in substantial degradation of diagnostic performance in new domain settings. Furthermore, the adoption of fine-tuning strategies, such as federated personalization, increases operational complexity, diminishes the practicality of FedDG, and delays diagnostic decisions in clinical environments, such as during sudden outbreaks or when integrating new hospitals.

Motivated by the aforementioned issues, we propose a novel tree-based personalized federated domain generalization framework (TreeFedDG), to address the FedDG-GD problem. First, during the model aggregation process, we design a hierarchical model aggregation mechanism. This mechanism deploys client models at the leaf nodes of a tree structure and performs bottom-up layered aggregation based on parameter similarity, effectively suppressing directional deviations in the global model. Simultaneously, we introduce a parameter difference-based style mixing method (FedStyle), which enforces style mixing among clients with maximum parameter differences, thereby enhancing the model’s robustness against parameter drift. Second, in the model dissemination process, we develop a progressive fusion mechanism. This partial aggregation strategy ensures the effective transmission of global knowledge while preserving the personalized features of each node. Additionally, the tree structure maintains a multi-granularity representation capability from leaf to root nodes by retaining the complete parameter space of each node. Unlike traditional federated learning approaches that retain only a single global model, our framework effectively leverages the specificity of client models and the advantages of the global model during data processing, thereby enhancing generalization to unseen domains. Finally, during the inference stage, we employ a feature similarity-guided model selection approach. This method utilizes cross-domain feature similarity to retrieve the most relevant model chain (from the matching leaf node to the root node path) within the tree structure for ensemble decision-making. 
 The main contributions of this paper are summarized as follows:
\begin{itemize}
    \item We propose a new and practical problem: the global drift issue in federated domain generalization for medical imaging (FedDG-GD).
    \item We introduce a simple and effective personalized federated domain generalization framework based on a tree-structured topology. This framework suppresses model drift through a hierarchical parameter aggregation mechanism and a parameter difference-based style mixing method (FedStyle); it incorporates a progressive model fusion strategy to achieve a dynamic balance between global knowledge and local features; and it develops a feature similarity-guided model selection mechanism that leverages the tree structure for client model retrieval, fully exploiting cross-domain feature similarities to enhance model generalization capabilities.
    \item Experiments conducted on two medical image segmentation datasets demonstrate that our proposed method outperforms advanced federated domain generalization approaches and achieves superior balance in cross-domain performance.
\end{itemize}

\begin{figure*}[ht]
  \centering
  \includegraphics[width=0.8\textwidth]{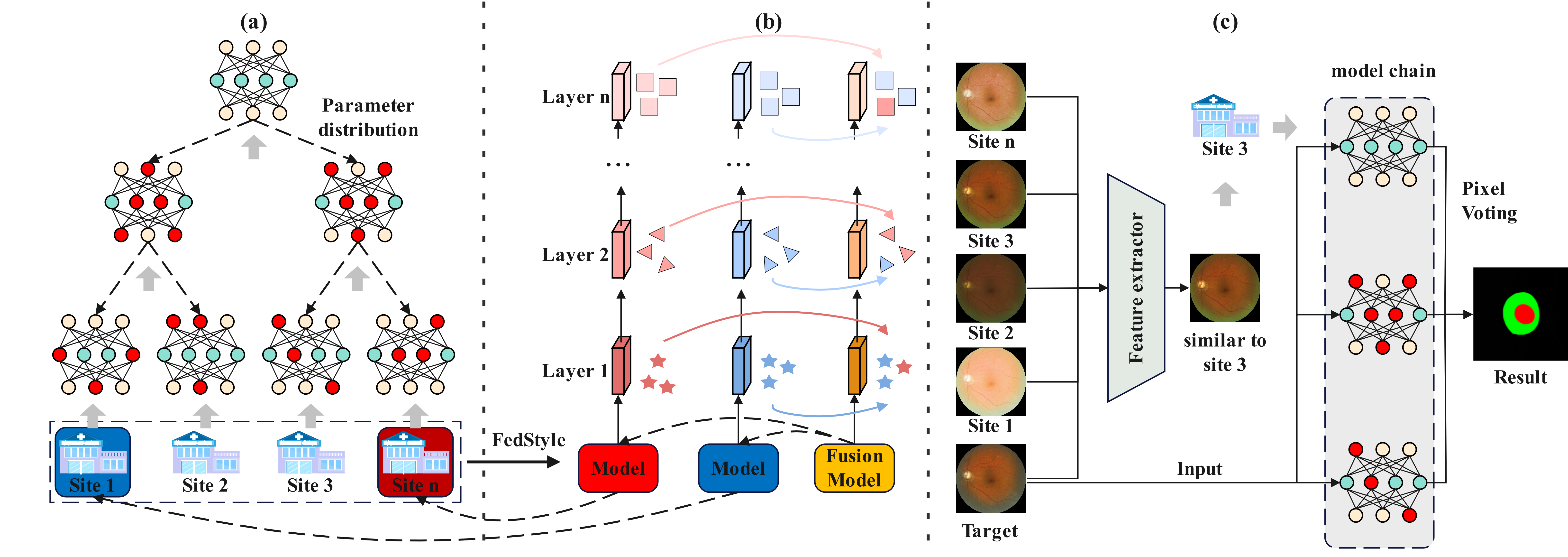}
  \caption{Overall framework workflow. (a) Hierarchical parameter aggregation and progressive personalization fusion of the tree-structured model. (b) Style mixing during client training. (c) Similarity-guided model selection during the inference phase.}
  \label{fig-method}
\end{figure*}


\section{Related work}
\subsubsection{Federated Domain Generalization in Medical Imaging}
With the widespread application of deep learning in medical image segmentation, achieving model generalization capability while protecting data privacy has become a key research focus. Federated Learning (FL) offers a distributed collaborative framework that, when combined with the high heterogeneity demands in medical imaging, has given rise to Federated Domain Generalization (FedDG) techniques \cite{lai2024bilateral}. FedDG aims to develop a global model with cross-domain generalization capabilities through federated training, enabling generalized segmentation performance without accessing samples from unseen target domains. For example, FedDG methods leverage local data privacy from various medical institutions and enhance domain generalization using techniques such as meta-learning \cite{liu2023domain}, style transfer \cite{chen2023federated}, feature alignment \cite{nguyen2022fedsr}, and frequency domain mixing \cite{pan2025frequency}. These approaches primarily focus on simulating or augmenting distribution diversity among source domains, or exploring strategies in local model training and feature space alignment \cite{park2023stablefdg,xu2023federated,wicaksana2022fedmix}.

However, existing FedDG methods generally emphasize local training and optimization of multi-source data, assuming that the parameter feature space can be effectively aligned and transferred during global aggregation. This theoretically simplifies the cross-domain generalization problem; however, in medical scenarios characterized by extremely high multi-source heterogeneity and complex data distributions, differences in devices, protocols, and patient populations across platforms exacerbate the global drift issue during parameter aggregation. Such directional deviation in global parameters after multiple aggregation rounds causes model performance to vary with source domain distributions, making it difficult to adapt to the complex characteristics of unseen domains.

\subsubsection{Drift Problem in Federated Learning}
The global drift of models is already a critical issue in traditional federated learning. To alleviate this problem, existing methods, such as personalized federated learning, have been proposed to enhance model adaptation on local data by designing client-specific feature spaces and probability densities \cite{liao2024foogd,song2024tackling,jiang2022harmofl}. Alternatively, normalization techniques and sample selection strategies have been introduced to suppress parameter shifts \cite{xu2025federated,kang2024fednn}. However, the primary goal of these strategies is to optimize the local model’s performance on its own (source domain) data, achieving “federated personalization” rather than “domain generalization” \cite{xu2022closing}. That is, existing personalization methods focus solely on the local optima for each client while neglecting the requirement for global model generalization to unseen domains. At the same time, personalization strategies often cause the client models to diverge gradually from the global model, preventing the global model from fully capturing multi-domain information and thus losing its generalization capability to new domains.

Furthermore, some cross-domain generalization methods attempt to mitigate drift by employing data augmentation, statistical alignment, and other measures \cite{cai2023fedce}. However, these approaches have two fundamental limitations: (1) they fail to address the deep-seated issue of random fluctuations in the direction of parameter aggregation; (2) they lack specialized designs for how the model should be selected or adapted in unseen domains. Consequently, existing methods for alleviating federated drift are difficult to directly migrate or extend to the FedDG-GD problem. In contrast, the tree-based topology structure proposed in this paper overcomes the limitations of traditional star-shaped architectures, providing a novel architectural paradigm for addressing the FedDG-GD problem in medical image analysis.

\section{Method}
\subsection{Definition and Overview}
In the federated learning domain generalization task, there exist source domains $D^S$ and an unknown target domain $D^T$, with each source domain $D^{S^i}$ corresponding to a federated learning client model $M_0^i$. Our objective is to construct a node tree $T$ through federated aggregation and utilize this tree to address the domain generalization problem. During training, $k$ source domains are used to train $k$ client models $M_0^1, M_0^2, \dots, M_0^k$. The tree structure is built through bottom-up layer-wise aggregation, followed by top-down parameter aggregation updates. After multiple iterations, a stable node tree $T$ is obtained. During inference, we first identify the client model $M_0^l$ corresponding to the source domain whose features are most similar to those of the target domain $D^T$ through feature comparison. Then, we trace upwards along the parent-child relationships in the tree structure to the root node, forming a complete model chain $C_l = \{M_0^l, M_1^l, M_2^l, \dots, M_H^l\}$, where $H$ is the maximum height of the node tree. Samples from the target domain $D^T$ are processed sequentially through each model in $C_l$, and the final segmentation result is output via pixel-wise weighted voting. The overall process is shown in Figure \ref{fig-method}.

\subsection{Hierarchical Parameter Aggregation}
To address the model drift issue in the federated training process, we design a hierarchical parameter aggregation method based on a tree-structured topology. This method constructs a node tree through model aggregation to preserve the models before and after aggregation along with their interrelationships for subsequent use. Algorithm \ref{alg1} describes the construction process of the node tree. Prior to federated aggregation, we first compute the cosine similarity \( S \) of parameters among models at the same layer and set a similarity threshold \( \tau \) for each layer. When the similarity \( S_{i,j} \leq \tau \) between two models \( M^i \) and \( M^j \), we group these two models into the same cluster. To ensure the rationality of clustering and the effectiveness of the resulting tree structure, we provide the following threshold setting method:

\begin{equation}
    \tau_l = \tau_0 + \beta \cdot \left( \frac{l}{H} \right)
\end{equation}

where \( \tau_l \) denotes the similarity threshold at layer \( l \), \( \tau_0 \) is the initial threshold value, \( H \) is the maximum height of the node tree, and \( \beta \) is the threshold adjustment coefficient that controls the variation amplitude of the similarity threshold from leaf nodes to the root node. This threshold setting approach emphasizes the similarity among clustered models, thereby effectively mitigating the model drift problem. Subsequently, starting from the leaf nodes, we perform federated aggregation based on the clustering results at each layer, storing the models involved in clustering and the cluster-generated models as parent-child nodes in the node tree.

\begin{algorithm}
    \caption{Hierarchical Parameter Aggregation}
    \label{alg1}
    \KwIn{$n$ clients, threshold $\tau_0$, max depth $H$}
    \KwOut{Node tree $T$}
        Initialize empty tree $T$
        \For{each training round}{
            Train leaf node models $M_0^i$ locally
            
            level $l \gets 1$
            
            \While{level $l$ has multiple models}{
                $\tau_l \gets \tau_0 + \beta \cdot (l / H)$ 
                
                Cluster models with similarity $\geq \tau_l$
                
                \For{each cluster}{
                    \If{cluster size $> 1$}{
                        Aggregate models and add to $l + 1$
                        
                        Connect as parent-child in $T$
                    }
                    \Else{
                        Promote single model to $l + 1$
                    }
                }
                $l \gets l + 1$
            }
            Update global model from tree root
        }
        \Return $T$
\end{algorithm}

\subsection{FedStyle}
Unlike existing methods, our proposed FedStyle method performs targeted style mixing based on parameter differences \cite{zhou2024mixstyle}. By identifying client pairs with the maximum differences in the parameter space, it facilitates style interactions that fully utilize the style diversity across different source domains while avoiding training noise from ineffective mixing. Specifically, before training the client model $M_0^i$, we identify the client model $M_0^j$ with the lowest parameter similarity to it, given by:
\begin{equation}
    j = \mathop{\mathrm{argmin}}_{j \neq i} \frac{\vec{\theta}_i \cdot \vec{\theta}_j}{\|\vec{\theta}_i\| \|\vec{\theta}_j\|}
\end{equation}

Where $\vec{\theta}_i$ and $\vec{\theta}_j$ represent the model parameters of client models $M_0^i$ and $M_0^j$, respectively. We then mix the style information from the source domain training data $D^{S_j}$ of client $j$ into the source domain training data $D^{S_i}$ of client $i$, resulting in the style-mixed source domain data $D^{S_{i,\text{Mix}}}$. To achieve this, we extract input feature map batches $x_i$ and $x_j$ from source domains $D^{S_i}$ and $D^{S_j}$, and compute the feature statistics $\mu$ and $\sigma$ as the style information to be mixed, where:

\begin{equation}
    \mu(x) = \frac{1}{HW} \sum_{h=1}^H \sum_{w=1}^W x_{h,w}
\end{equation}

\begin{equation}
    \sigma(x) = \sqrt{\frac{1}{HW} \sum_{h=1}^H \sum_{w=1}^W (x_{h,w} - \mu(x))^2}
\end{equation}

The mixed feature statistics are then calculated as:

\begin{equation}
    \beta_{\text{mix}} = \lambda \mu(x_i) + (1 - \lambda) \mu(x_j)
\end{equation}

\begin{equation}
    \gamma_{\text{mix}} = \lambda \sigma(x_i) + (1 - \lambda) \sigma(x_j)
\end{equation}

Where $\lambda \in \mathbb{R}$ is randomly sampled from a Beta distribution, i.e., $\lambda \sim \text{Beta}(\phi, \phi)$, with $\phi \in (0, +\infty)$ being a hyperparameter. We set $\phi$ to 0.1. Finally, the mixed feature statistics are applied to the style-normalized $x_i$ as follows:

\begin{equation}
    \text{FedStyle}(x_i) = \gamma_{\text{mix}} \cdot \frac{(x_i - \mu(x_i))}{\sigma(x_i)} + \beta_{\text{mix}}
\end{equation}

In practice, we activate the FedStyle module with a probability of 0.5 during forward propagation. During testing, the FedStyle module is not applied. Notably, the gradients for $\sigma(\cdot)$ and $\mu(\cdot)$ are blocked in the computation graph.

\subsection{Progressive Personalized Model Fusion}
In our work, models at different layers of the tree-structured topology learn global knowledge to varying degrees during the aggregation process while preserving their respective model specificities. To achieve this, we design a progressive personalized fusion strategy based on hierarchical parameter partitioning. Unlike traditional federated learning, which simply disseminates a global model, our approach employs hierarchical parameter aggregation to ensure effective transmission of global knowledge while maintaining the personalized characteristics of each node in the tree structure, thereby achieving an optimal balance between model generality and specificity.

Specifically, after each round of hierarchical parameter aggregation, TreeFedDG adopts a top-down parameter dissemination strategy. We divide each model parameter $\theta$ into two parts: fixed-layer parameters $\theta^{\text{fixed}}$ and variable-layer parameters $\theta^{\text{var}}$, such that $\theta = [\theta^{\text{fixed}}, \theta^{\text{var}}]$. Here, the fixed-layer parameters remain unchanged during dissemination to preserve model specificity, while the variable-layer parameters are updated through a progressive fusion mechanism to integrate global knowledge from upper-layer models.

When the model parameters of parent node $i$, denoted as $\theta_i = [\theta_i^{\text{fixed}}, \theta_i^{\text{var}}]$, are disseminated to child node $j$, the update strategy for child node $j$ is as follows:

\begin{equation}
    \theta_j^{\text{new}} = [\theta_j^{\text{fixed}}, \theta_j^{\text{var,new}}]
\end{equation}

The variable-layer parameters are fused progressively according to:

\begin{equation}
    \theta_j^{\text{var,new}} = \varepsilon_j \cdot \theta_i^{\text{var}} + (1 - \varepsilon_j) \cdot \theta_j^{\text{var,old}}
\end{equation}

Where $\theta_j^{\text{var,old}}$ is the value of the variable-layer parameters before the update for child node $j$, $\theta_j^{\text{var,new}}$ is the updated value, and $\varepsilon_j$ is the progressive fusion coefficient for node $j$ at layer $l$, defined as:

\begin{equation}
    \varepsilon_j = \varepsilon_0 \cdot \omega^{1 - l}, \quad \omega \in (0, 1)
\end{equation}

Where, $\varepsilon_0$ is the initial progressive fusion coefficient, and $\omega$ is the decay factor. The decay factor ensures that the fusion coefficient decreases as the tree structure approaches the leaf nodes, allowing leaf nodes to retain more local specificity while upper nodes incorporate more global knowledge.

\subsection{Feature Similarity-Guided Model Selection}
In our work, after a series of training processes, TreeFedDG constructs a tree-structured topology that encapsulates multi-granularity domain knowledge. Leaf nodes retain fine-grained features specific to particular domains, intermediate nodes fuse common knowledge from similar domains, and the root node carries global universal features. During the inference stage, it is necessary to select the optimal model combination for the unseen target domain to leverage this hierarchical knowledge effectively. Unlike traditional federated learning that uses only a single global model for inference, TreeFedDG employs a feature similarity-guided mechanism to retrieve the most matching model chain from the tree-structured topology. Specifically, we use a pre-trained general feature extractor \( f_{\text{enc}}(\cdot) \) to extract image features from each source domain and the target domain as follows:

\begin{equation}
    \phi(x) = \text{Agg}(f_{\text{enc}}(x))
\end{equation}

Where \( x \) is an image from various domains, and \( \text{Agg} \) is the feature aggregation function, defined as:

\begin{equation}
     \text{Agg}(f) = [\mu(f), \sigma(f), \text{Hist}(f)] 
\end{equation}

Where, \( \mu(f) \) and \( \sigma(f) \) represent the mean and standard deviation of the features, respectively, and \( \text{Hist}(f) \) is the histogram statistic of the features. This allows us to obtain the overall image features for each source domain \( \Phi(D^{S^i}) = \{\phi(x) | x \in D^{S^i}\} \) and for the target domain \( \Phi(D^T) = \{\phi(x) | x \in D^T\} \). By comparing the feature similarity between the target domain and the source domains, we identify the source domain most similar to the target domain:

\begin{equation}
    D^{S_i} = \arg\max_{D^{S^i}} \text{sim}(\Phi(D^{S^i}), \Phi(D^T))
\end{equation}

Guided by this source domain, we locate its corresponding client model \( M_0^l \), and subsequently find all parent node models along the tree structure, forming a model chain \( C_l = \{M_0^l, M_1^l, M_2^l, \ldots, M_H^l\} \). For test samples, we adopt an ensemble decision strategy based on weighted pixel-wise voting. Specifically, for a segmentation task, each model in the chain produces a class prediction for each pixel position \( (u, v) \). We assign different weights based on the model's level in the chain and feature similarity, and then perform pixel-level weighted voting:

\begin{equation}
    \hat{y}_t(u, v) = \arg\max_c \sum_{M_h^l \in C_l} w_h \cdot \mathbf{1}[M_h^l(x_t)(u, v) = c]
\end{equation}

Where \( \hat{y}_t(u, v) \) is the final predicted class at position \( (u, v) \), \( \mathbf{1}[\cdot] \) is the indicator function that equals 1 if the prediction is class \( c \), and 0 otherwise. The weight \( w_h \) for models at level \( h \) is defined as:

\begin{equation}
    w_h = \frac{\exp(-\partial \cdot h)}{\sum_{h' = 0}^H \exp(-\partial \cdot h')}
\end{equation}

In this formula, \( \partial \) is a hyperparameter controlling the importance of levels, and the term \( \exp(-\partial \cdot h) \) assigns higher weights to models closer to the leaf nodes (lower levels) because they retain more domain-specific information.

\begin{table*}[]
	\centering
	\begin{adjustbox}{width=\textwidth}     
	\begin{tabular}{c|ccccc|ccccc|c||cccccccccc|c}
		\hline \hline
		Task        & \multicolumn{5}{c|}{Optic Disc Segmentation}              & \multicolumn{5}{c|}{Optic Cup Segmentation}               & \multirow{2}{*}{Overall} & \multicolumn{5}{c}{Optic Disc Segmentation}               & \multicolumn{5}{|c}{Optic Cup Segmentation}               & \multicolumn{1}{|c}{\multirow{2}{*}{Overall}} \\ \cline{1-11} \cline{13-22}
		Unseen Site & A         & B         & C         & D         & Avg       & A         & B         & C         & D         & Avg       &                          & A         & B         & C         & D         & Avg       & \multicolumn{1}{|c}{A} & B         & C         & D         & Avg       &                                               \\ \hline \hline
		            & \multicolumn{11}{c|}{Dice Coefficient (Dice) ↑}                                                                                                                                                                                                                                                        & \multicolumn{11}{c}{Hausdorff Distance (HD95) ↓}                                                                                                                                                                                                                                                                                                                                                                                                                                                                                                \\ \hline \hline
		FedAvg \cite{mcmahan2017communication}     & 88.77     & 79.19     & 87.44     & 87.7      & 85.78     & 75.64     & 63.07     & 75.52     & 77.23     & 72.87     & 79.33                    & 16.42     & 23.58     & 15.89     & 11.76     & 16.91     & 24.72                  & 25.43     & 17.28     & 13.57     & 20.25     & 18.58                                         \\
		FedCE \cite{cai2023fedce}      & 90.33     & 81.35     & 88.91     & 89.33     & 87.48     & 77.81     & 65.3      & 77.69     & 79.38     & 75.05     & 81.27                    & 14.35     & 21.36     & 14.00        & 10.44     & 15.04     & 22.08                  & 23.21     & 15.36     & 12.43     & 18.27     & 16.66                                         \\
		FedGA \cite{zhang2023federated}      & 92.67     & 84.59     & 91.12     & 91.78     & 90.04     & 81.07     & 68.65     & 80.95     & 82.61     & 78.32     & 84.18                    & 12.97     & 19.88     & 12.74     & 9.56      & 13.79     & 20.32                  & 21.73     & 14.08     & 11.67     & 16.95     & 15.37                                         \\
		L-DAWA \cite{rehman2023dawa}     & 91.89     & 83.51     & 90.38     & 90.96     & 89.19     & 79.98     & 67.53     & 79.86     & 81.53     & 77.23     & 83.21                    & 15.04     & 22.1      & 14.63     & 10.88     & 15.66     & 22.96                  & 23.95     & 16.00        & 12.81     & 18.93     & 17.3                                          \\
		FedDG \cite{liu2021feddg}      & 92.31     & 84.12     & 90.78     & 91.39     & 89.65     & 80.54     & 68.12     & 80.43     & 82.09     & 77.8      & 83.72                    & 14.27     & 21.42     & 13.95     & 10.51     & 15.04     & 22.13                  & 23.17     & 15.29     & 12.38     & 18.24     & 16.64                                         \\
		FedUAA \cite{wang2023federated}     & 94.23     & 86.75     & 92.58     & 93.41     & 91.74     & 83.24     & 70.88     & 83.12     & 84.76     & 80.5      & 86.12                    & 13.66     & 20.62     & 13.37     & 10        & 14.41     & 21.2                   & 22.47     & 14.72     & 12.05     & 17.61     & 16.01                                         \\
		FedEvi \cite{chen2024fedevi}     & 95.12     & 87.84     & \textbf{93.65}     & 94.23     & 92.71     & 84.45     & 72.15     & 84.21     & 81.83     & 80.66     & 86.69                    & 12.97     & 19.88     & 12.74     & 9.56      & 13.79     & 20.32                  & 21.73     & 14.08     & 11.67     & 16.95     & 15.37                                         \\ \hline
		Ours        & \textbf{95.27}     & \textbf{90.68}     & 93.26     & \textbf{94.34}     & \textbf{93.38}     & \textbf{86.12}     & \textbf{78.57}     & \textbf{85.71}     & \textbf{82.36}     & \textbf{83.19}     & \textbf{88.29}                    & \textbf{11.24}     & \textbf{15.32}     & \textbf{10.76 }    & \textbf{8.32}      & \textbf{11.41}     & \textbf{17.56 }                 & \textbf{17.95}     & \textbf{13.32}     & \textbf{9.81}      & \textbf{14.66}     & \textbf{13.03}                                         \\ \hline \hline
	\end{tabular}
	\end{adjustbox}
	\caption{Performance comparison results on the fundus image segmentation dataset}
	\label{tab-funds}
\end{table*}

\begin{table*}[]
	\centering
	\begin{adjustbox}{width=0.8\textwidth}     
	\begin{tabular}{c|ccccccc|ccccccc}
		\hline \hline
		Unseen Site & A       & B       & C       & D       & E       & F       & Avg     & A       & B       & C       & D       & E       & F       & Avg     \\ \hline \hline
		            & \multicolumn{7}{c|}{Dice Coefficient (Dice) ↑}                      & \multicolumn{7}{c}{Hausdorff Distance (HD95) ↓}                       \\ \hline \hline
		FedAvg \cite{mcmahan2017communication}     & 80.25   & 88.26   & 90.69   & 83.16   & 87.07   & 85.17   & 85.77   & 16.12   & 9.64    & 5.8     & 14.06   & 10.17   & 8.6     & 10.73   \\
		FedCE \cite{cai2023fedce}      & 83.31   & 88.71   & 91.55   & 84.18   & 88.4    & 82.71   & 86.48   & 12.95   & 9.32    & 5.35    & 13.9    & 8.56    & 8.67    & 11.79   \\
		FedGA \cite{zhang2023federated}      & 85.84   & 87.86   & 89.91   & 81.26   & 84.93   & 83.61   & 85.57   & 11.59   & 10.29   & 6.64    & 16.36   & 11.77   & 11.85   & 11.42   \\
		L-DAWA \cite{rehman2023dawa}     & 83.94   & 87.79   & 89.66   & 79.92   & 87.03   & 86.55   & 85.82   & 10.15   & 11.59   & 11.94   & 10.45   & 13.76   & 9.28    & 11.2    \\
		FedDG \cite{liu2021feddg}      & 84.68   & 88.25   & 90.51   & 81.08   & 87.54   & 85.37   & 86.24   & 11.12   & 10.38   & 8.74    & 13.18   & 11.31   & 9.29    & 10.67   \\
		FedUAA \cite{wang2023federated}     & 85.53   & 88.77   & 91.24   & 82.28   & 88.15   & 84.1    & 86.68   & 11.95   & 9.25    & 5.46    & 15.95   & 8.77    & 9.31    & 10.12   \\
		FedEvi \cite{chen2024fedevi}     & 87.09   & 88.52   & 90.23   & 88.4    & 89.11   & 85.25   & 88.1    & 9.61    & 9.39    & 6.02    & 10.01   & 7.65    & 8.35    & 8.51    \\ \hline
		Ours        & \textbf{89.13}   & \textbf{90.62}   & \textbf{92.32}   & \textbf{90.29}   & \textbf{91.83}   & \textbf{87.94}   & \textbf{90.35}   & \textbf{9.03}    & \textbf{8.85}    & \textbf{5.12}    & \textbf{9.47}    & \textbf{6.54}    & \textbf{8.12}    & \textbf{7.43}    \\ \hline \hline
	\end{tabular}
	\end{adjustbox}
	\caption{Performance comparison results on the prostate MRI segmentation dataset}
	\label{tab-prostate}
\end{table*}

\section{Experiments and Results}
\subsection{Experimental Setting}
\subsubsection{Datasets and Preprocessing}
We evaluate the proposed method on two widely used medical image segmentation datasets: the Fundus dataset for retinal image segmentation \cite{fumero2011rim,orlando2020refuge,sivaswamy2015comprehensive} and the Prostate dataset for multi-site prostate MRI segmentation \cite{lemaˆitrecomputer,litjens2014evaluation,liu2020ms}. We adopt a leave-one-domain-out evaluation protocol, where in each trial, one domain is selected as the unseen target domain (test domain), and all remaining domains are used as source domains (training domains). The training and validation set splits within each source domain are consistent with those in \cite{mcmahan2017communication,cai2023fedce,zhang2023federated}, and the entire target domain is used for testing.

\subsubsection{Implementation Details}
All experiments were conducted in a Python 3.9 and PyTorch 2.4 environment on an Ubuntu 22.04 system, utilizing 5 NVIDIA RTX 4090 GPUs. In the federated learning process, all clients used the same hyperparameter settings. We adopted a lightweight U-Net as the base segmentation network architecture. During local training, the batch size was set to 8, the learning rate to \(1 \times 10^{-4}\), and the number of local epochs \(E\) to 50, with a total of 100 communication rounds \(R\). For TreeFedDG, the initial similarity threshold \(\tau_0\) was set to 0.85, the maximum tree depth \(D_{\text{max}}\) to 3, the initial progressive fusion coefficient \(\varepsilon_0\) to 0.8, the decay factor \(\omega\) to 0.5, the FedStyle mixing probability to 0.5, and the model hierarchy weight hyperparameter \(\beta\) to 0.5. For feature similarity computation, we used a pretrained ResNet-18 as the feature extractor, with its parameters frozen during training to only extract image features. Evaluation metrics included the Dice coefficient (Dice) and the 95\% Hausdorff Distance (HD95).

\subsection{Results and Comparative Analysis}
We compare TreeFedDG with existing federated learning and domain generalization methods to verify its effectiveness on medical image segmentation tasks. 

\subsubsection{The comparison results of performance} Table \ref{tab-funds} presents a performance comparison of different methods on the retinal image segmentation dataset. Specifically, in terms of the Dice coefficient metric, our method achieves an average of 88.29, and for the HD95 metric, our method has an average of only 13.03, both surpassing the existing SOTA methods. The experimental results on the prostate MRI segmentation task further validate the superior performance of the TreeFedDG framework (as shown in Table \ref{tab-prostate}). Our method again achieves the best overall performance, with an average Dice coefficient of 90.35 and an average HD95 metric of only 7.43, significantly outperforming SOTA methods. This highlights the superiority of TreeFedDG in improving segmentation accuracy and reducing boundary errors, demonstrating its strong generalization ability in handling domain shift issues in medical images.

\begin{figure}[ht]
  \centering
  \includegraphics[width=\linewidth]{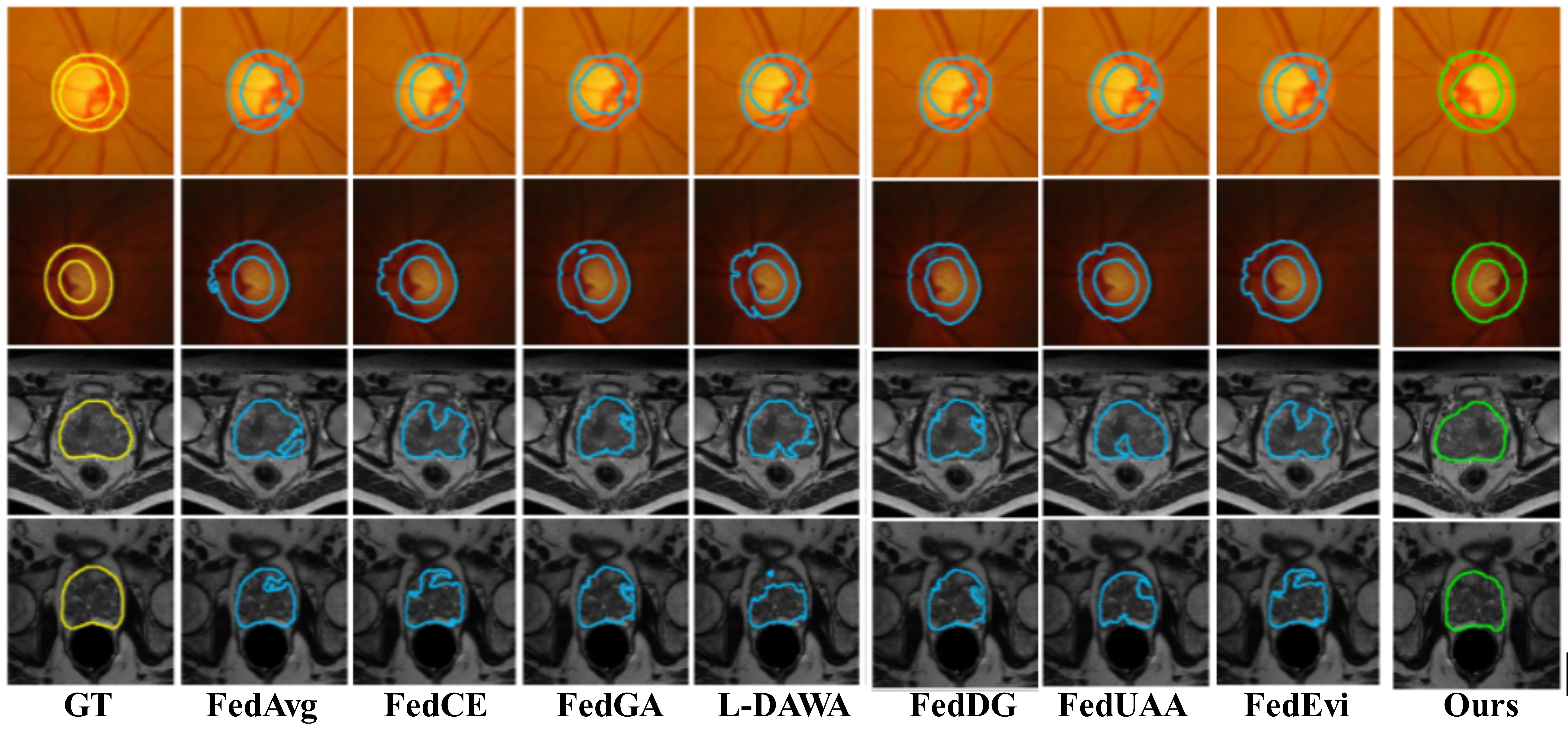}
  \caption{Qualitative comparison of the generalization results of different methods for fundus image segmentation (the top two rows) and prostate MRI segmentation (the bottom two rows)}
  \label{fig-result}
\end{figure}

To visually demonstrate the advantages of TreeFedDG, Figure \ref{fig-result} shows the visualization results of different methods on the retinal image segmentation task. It can be observed that traditional federated learning methods often suffer from blurry boundaries or incomplete segmentation when handling target domain images. Existing federated domain generalization methods show some improvements but still exhibit errors in handling detailed structures. In contrast, TreeFedDG accurately identifies the boundaries of the optic cup and optic disc, with segmentation results closely matching the ground truth annotations.

\subsubsection{The comparison results of performance consistency} 
\begin{figure}[ht]
  \centering
  \includegraphics[width=\linewidth]{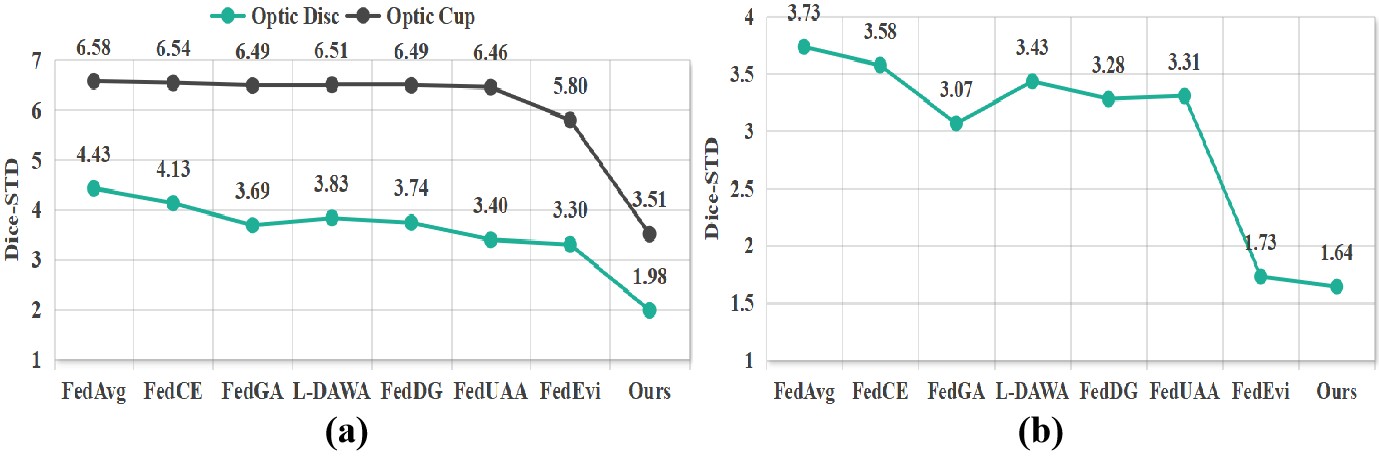}
  \caption{Qualitative comparison of Dice performance consistency across different methods for retinal image segmentation and prostate MRI segmentation. (Lower STD values indicate smaller performance variations and stronger model generalization capabilities.)}
  \label{fig-std}
\end{figure}

To assess the effectiveness of TreeFedDG in addressing global drift issues in federated domain generalization tasks, we focus on the model’s performance consistency across different sites. To quantify this issue, we use the standard deviation (STD) as a metric to calculate the performance fluctuation in Dice coefficients on unseen sites. The experimental results are shown in Figure \ref{fig-std}.

Experimental results show that on the retinal image segmentation dataset (Figure \ref{fig-std}.a), TreeFedDG achieves a Dice STD of 1.98 for Optic Disc segmentation and 3.51 for Optic Cup segmentation, which is significantly lower compared to existing methods. This indicates that TreeFedDG provides more balanced performance across different domains, reducing performance fluctuations even in the complex Optic Cup segmentation task. Similarly, on the prostate segmentation dataset (Figure \ref{fig-std}.b), TreeFedDG has a Dice STD of only 1.64, which is substantially lower than other methods. This further confirms that TreeFedDG effectively alleviates the FedDG-GD problem, avoiding the large discrepancies observed in existing methods at specific sites.

\subsection{Ablation Study}
To comprehensively verify the effectiveness of the proposed TreeFedDG framework, we conduct a series of ablation studies, analyzing the core components of the framework from four key aspects: (1) the effectiveness of the tree-structured topology; (2) the impact of FedStyle; (3) the impact of the progressive fusion mechanism; and (4) the contribution of the feature similarity-guided model selection. All ablation experiments are performed on the retinal image segmentation dataset, with the Dice coefficient serving as the primary evaluation metric.

\begin{figure}[ht]
  \centering
  \includegraphics[width=\linewidth]{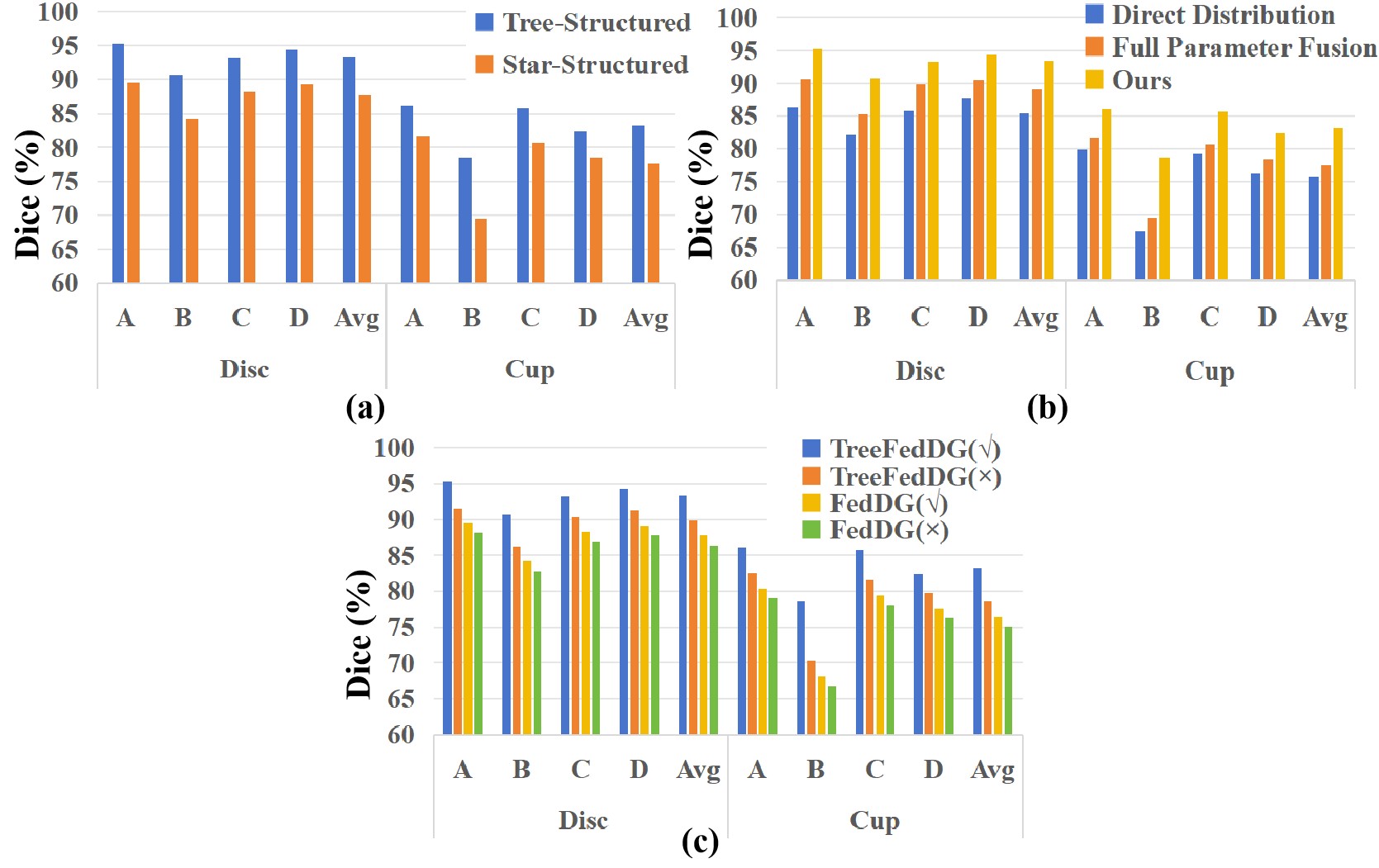}
  \caption{Ablation study results on (a) the effectiveness of the tree-structured architecture, (b) the impact of the FedStyle method, and (c) the impact of the progressive fusion mechanism.}
  \label{fig-ab1}
\end{figure}

\subsubsection{Effectiveness of Tree-Structured Topology}
We first analyze the advantages of the tree-structured topology in TreeFedDG compared to the traditional star-structured topology in federated learning. As shown in Figure \ref{fig-ab1}.a, we keep all other components unchanged and only modify the network topology from a tree structure to a star structure (i.e., the central node is directly connected to all client nodes in the traditional federated learning setup). Experimental results indicate that the tree-structured topology outperforms the star-structured topology across all target domains.

\subsubsection{Impact of the FedStyle Method}
To verify the effectiveness of the FedStyle method in addressing model drift issues, we conducted comparative experiments on tree-structured topologies with and without the FedStyle method. The results are presented in Figure \ref{fig-ab1}.b. Experimental results show that FedStyle brings performance improvements on both network topology structures, but the enhancement is more significant in the tree-structured topology. This indicates that FedStyle, by enforcing style mixing among clients with the largest parameter differences, effectively enhances the model’s robustness against parameter drift.

\subsubsection{Impact of the Progressive Fusion Mechanism}
Our proposed progressive fusion mechanism is crucial for maintaining model specificity and facilitating global knowledge transfer. Under the tree-structured topology, we compare three different strategies: (1) direct distribution (where the parent node directly overwrites the child node’s parameters); (2) full parameter fusion (where all parameters of the parent node are fused with those of the child node); (3) layered progressive distribution (our method, where only partial parameters of the parent node are fused with the child node). The comparison results are shown in Figure \ref{fig-ab1}.c. Experimental results indicate that the layered progressive distribution strategy achieves the best performance.

\subsubsection{Contribution of the Feature Similarity-guided Model Selection}
\begin{figure}[ht]
  \centering
  \includegraphics[width=\linewidth]{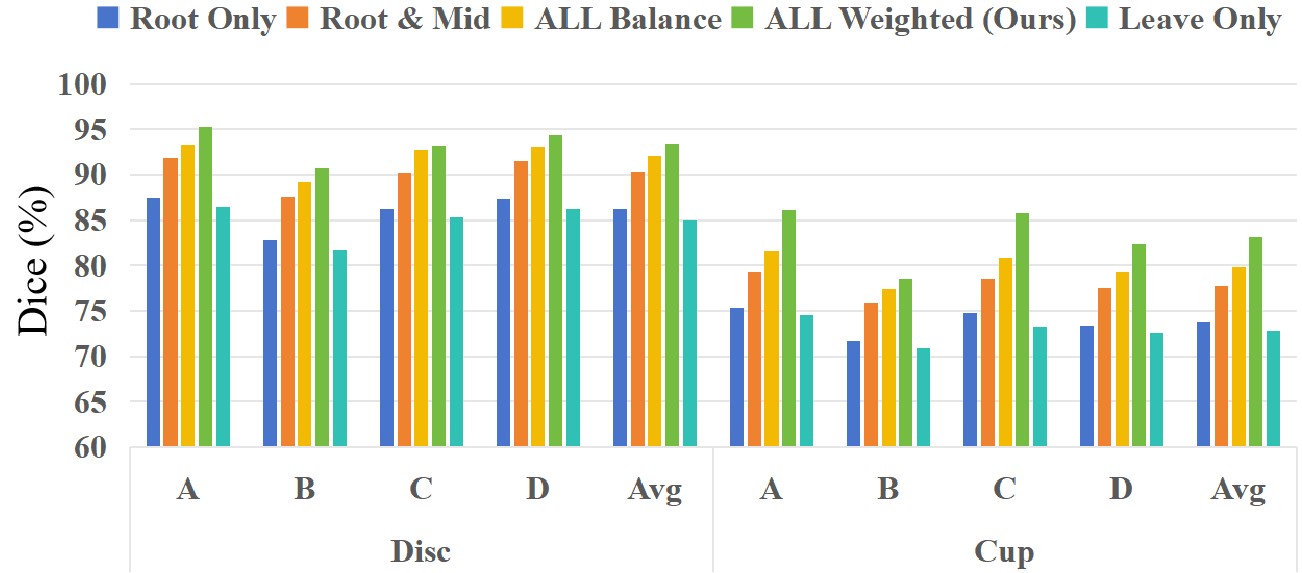}
  \caption{Ablation study results on the feature similarity-guided model selection strategy.}
  \label{fig-ab2}
\end{figure}

Finally, we evaluate the effectiveness of the tree-based tracing voting selection strategy during the inference phase. We compare five different model selection strategies: (1) using only the root node model; (2) using only the root and intermediate node models (excluding leaf nodes); (3) using models from all nodes with equal weights; (4) using models from all nodes with weights assigned based on hierarchy levels (our method); (5) using only the single most similar leaf node model. The results are presented in Figure \ref{fig-ab2}. Experimental results show that the full-tree model selection strategy with hierarchy-based weighting achieves the best performance.

Overall, the ablation study results validate the effectiveness and necessity of each component in the TreeFedDG framework. The tree-structured topology effectively suppresses model drift through hierarchical aggregation; the FedStyle style mixing enhances model robustness; the progressive fusion mechanism achieves effective global knowledge transfer while maintaining model specificity; and the feature similarity-guided model selection strategy leverages cross-domain feature similarities to improve model generalization on unseen domains.
\section{Conclusion}
This paper first defines and addresses the global drift problem in federated domain generalization (FedDG-GD). To tackle this issue, we propose TreeFedDG, a novel framework with tree-structured topology, which aggregates parameters based on dissimilarity, balances global and local knowledge via progressive fusion, and uses feature similarity for inference to boost generalization. Experiments show it surpasses SOTA methods on public datasets with superior cross-domain consistency.
\bibliography{aaai2026}

\end{document}